\documentclass{article}

\PassOptionsToPackage{numbers}{natbib}

\usepackage[preprint]{neurips_2023}

\usepackage[utf8]{inputenc} 
\usepackage[T1]{fontenc}    
\usepackage{hyperref}       
\usepackage{url}            
\usepackage{booktabs}       
\usepackage{amsfonts}       
\usepackage{nicefrac}       
\usepackage{microtype}      
\usepackage{xcolor}         
\usepackage{soul}
\usepackage{graphicx}
\usepackage{verbatim}
\usepackage{caption}
\usepackage{subcaption}
\usepackage{footmisc}
\usepackage{booktabs}
\usepackage{hhline}
\usepackage{multirow}
\usepackage{float}
\usepackage{xltabular}
\usepackage{lmodern}
\usepackage[most]{tcolorbox}
\tcbuselibrary{skins}
\usepackage{algpseudocode}
\usepackage{longtable}
\usepackage{framed}
\usepackage{hyperref}
\usepackage{enumitem}  
\usepackage{adjustbox}
\usepackage{makecell}
\usepackage{multirow}
\usepackage{booktabs}
\usepackage{amsmath}  
\usepackage{tcolorbox} 
\usepackage{mdframed}

\tcbset{
  aibox/.style={
    width=\textwidth,
    top=0pt, bottom=0pt, left=5pt, right=5pt,
    colback=white,
    colframe=black,
    colbacktitle=black,
    enhanced,
    center,
    attach boxed title to top left={yshift=-0.1in,xshift=0.15in},
    boxed title style={boxrule=0pt,colframe=white,},
  }
}
\newtcolorbox{AIbox}[2][]{aibox,title=#2,#1}

\usepackage[ruled,vlined]{algorithm2e}
\newcommand{\squishlist}{
   \begin{list}{$\bullet$}
    { \setlength{\itemsep}{0pt}      \setlength{\parsep}{3pt}
      \setlength{\topsep}{3pt}       \setlength{\partopsep}{0pt}
      \setlength{\leftmargin}{1.5em} \setlength{\labelwidth}{1em}
      \setlength{\labelsep}{0.5em} } }

\newcommand{\squishlisttwo}{
   \begin{list}{$\bullet$}
    { \setlength{\itemsep}{0pt}    \setlength{\parsep}{0pt}
      \setlength{\topsep}{0pt}     \setlength{\partopsep}{0pt}
      \setlength{\leftmargin}{2em} \setlength{\labelwidth}{1.5em}
      \setlength{\labelsep}{0.5em} } }

\newcommand{\squishend}{
    \end{list}  }

\usepackage{blindtext}

\newtcolorbox[list inside=mybox,auto counter,number within=section]{MyBox}{colbacktitle=yellow,coltitle=black,title={MyBox \thetcbcounter}}

\usepackage{tikz}

\newcommand{\noborderthanks}[1]{%
    \begingroup  
    \hypersetup{pdfborder={0 0 0}}%
    \thanks{#1}%
    \endgroup  
}

\title{%

\raisebox{-0.3cm}{\includegraphics[width=1cm, height=1cm]{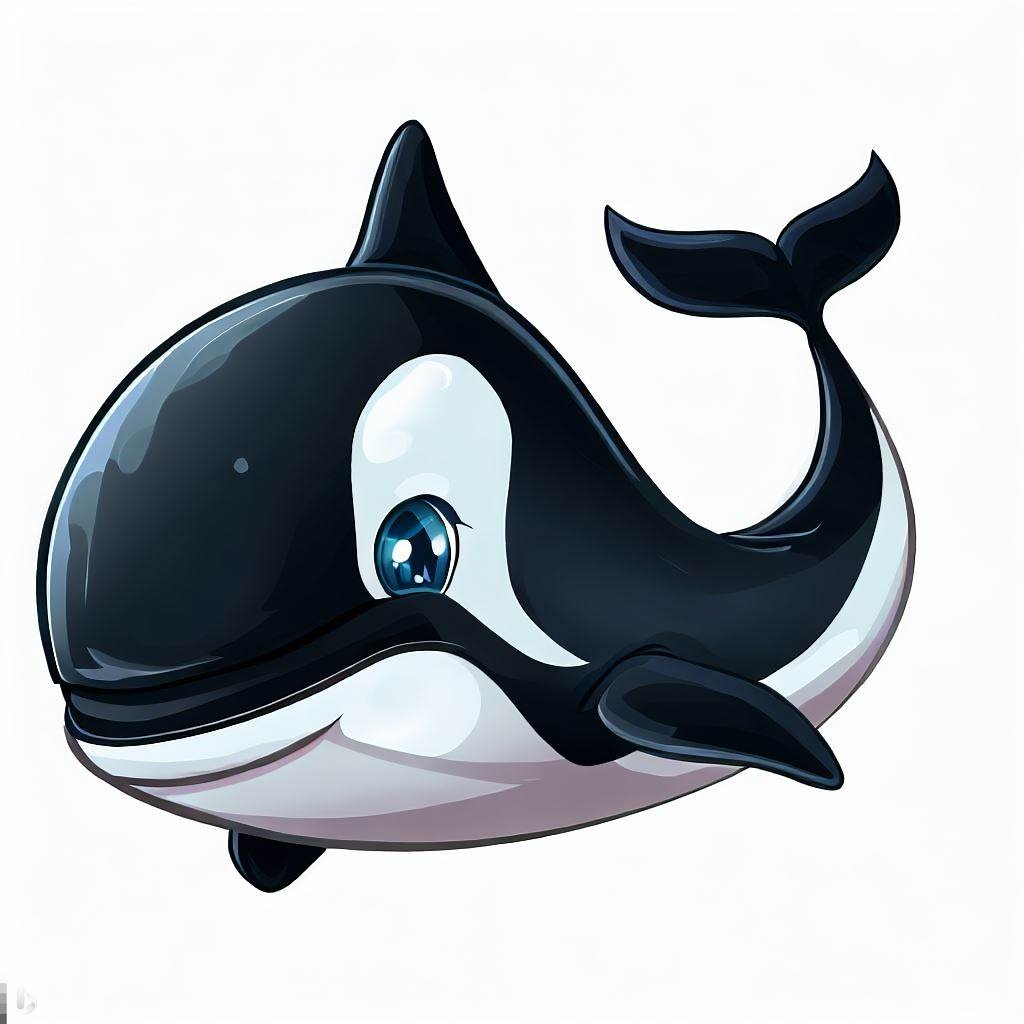}}\ {Orca-Math: Unlocking the potential of SLMs in Grade School Math }

}

\author{Arindam Mitra\noborderthanks{Correspondence to \texttt{arindam.mitra@microsoft.com}},
{\bf  Hamed Khanpour, Corby Rosset, Ahmed Awadallah} \AND Microsoft Research\vspace{0em}}

\begin{document}

\maketitle
\begin{abstract}

We show that an SLM can reach $\sim 87\%$ pass@1 on GSM8K  while trained on only 200K synthetic math problems.
Mathematical word problem-solving has long been recognized as a complex task for small language models (SLMs). A recent study hypothesized that the smallest model size, needed to achieve over 80\% accuracy on the GSM8K benchmark, is 34 billion parameters. To reach this level of performance with smaller models, researcher often train SLMs to generate Python code or use tools to help avoid calculation errors. Additionally, they employ ensembling, where outputs of up to 100 model runs are combined to arrive at a more accurate result. Result selection is done using consensus, majority vote or a separate a verifier model used in conjunction with the SLM. Ensembling provides a substantial boost in accuracy but at a significant cost increase with multiple calls to the model (e.g., Phi-GSM uses top-48 to boost the performance from  68.2 to 81.5, ~\cite{OVM} uses top-100 to boost LLAMA-2's performance from 38.6\% to 71.9\%).\\

In this work, we present Orca-Math, a 7-billion-parameter SLM based on the Mistral-7B, which achieves 86.81\% on GSM8k without the need for multiple model calls or the use of verifiers, code execution or any other external tools. Our approach has the following key elements: (1) A high quality synthetic dataset of 200K math problems created using a multi-agent setup where agents collaborate to create the data, (2) An iterative learning techniques that enables the SLM to practice solving problems, receive feedback on its solutions and learn from preference pairs incorporating the SLM solutions and the feedback.
When trained with Supervised Fine-Tuning alone, Orca-Math achieves
81.50\% on GSM8k pass@1 metric. With iterative preference learning, Orca-Math achieves 86.81\% pass@1. Orca-Math surpasses the performance of significantly larger models such as LLAMA-2-70B, WizardMath-70B, Gemini-Pro, ChatGPT-3.5. It also significantly outperforms other smaller models while using much smaller data (hundreds of thousands vs. millions of problems).


\end{abstract}

\clearpage
\section{Problem Setup}

Frontier Language Models such as GPT-4 \cite{achiam2023gpt} have demonstrated capabilities previously unseen in smaller models, most notably the remarkable ability to reason (e.g. mathematical reasoning that requires both language comprehension and mathematical understanding). These capabilities have been largely attributed to the very large scale the model size, the dataset size and ultimately the amount of compute needed for training.

Several recent studies have focused on improved the reasoning abilities of small language models (SLMs). Despite that the extent to which scale is needed for achieving reasoning capabilities is still an open research question.

One of the promising directions of improving the reasoning capabilities of SLMs is using frontier language models, such as GPT-4, to create tailored and high-quality synthetic data that can be used to train the SLM. The high quality of the training data and the ability to elicit richer learning signals (e.g. explanations) have been show to significantly improve SLMs abilities in acquiring skills that had only emerged before at much larger scale.

This paradigm fits under a teacher-student approach where the large model (the teacher) is creating demonstrations for the SLM (the student) to learn from. In this work we further explore this direction with focus on mathematical reasoning on grade school math world problem, using the popular GSM8K benchmark.

Several other studies have demonstrated positive results on GSM8K recently with SLMs, e.g. Phi-GSM \cite{liu2023tinygsm}, OVM \citep{OVM}, etc. However, many of them employ ensembling, where outputs of up to 100 model runs are combined to arrive at a more accurate results. Result selection is done using, consensus, majority vote or by using a separate a verifier model to score/verify the outputs and select the best answer. Ensembling provides a substantial boost in accuracy  (e.g., Phi-GSM uses top-48 to boost the performance from 68.2 to 81.5, [22] uses top-100 to boost LLAMA-2’s performance from 38.6\% to 71.9\%). However it comes at a significant increase in cost with multiple calls to the model, generating and verifying a 100 different solutions requires 200 different calls to the models. Additionally, some of them use very larger amounts of data (e.g. 12M for Phi-GSM) or use tools or code to avoid calculation errors.

In this work, we extend the teacher-student paradigm to an iterative learning settings with high-quality synthetic training data as follows:
\begin{itemize}
\item We create Orca-Math-dataset, a synthetic dataset of 200K math problems, paired with GPT-4-Turbo solutions. The dataset was generated using an agent-based setup, hereby referred as, Agent-Instruct, that not only paraphrases existing problems but aims to expand the problem set both in diversity and difficulty. 
\item We introduce an iterative learning procedure where we: (1) use the dataset for supervised finetuning to train the SLM on demonstrations, (2) allow the SLM to practice generating multiple solutions and (3) use the teacher to provide feedback to the student. The feedback comes in the form of evaluating the solutions generated by the student or providing a teacher solution. 
\end{itemize}

With the supervised finetuning alone, we achieve 81.50\% on GSM8k at pass@1 metric. The iterative learning loop further improves the pass@1 to 86.81\%. without the need for multiple model calls or the use of verifiers, code execution or any other external tools. The model exceeding much bigger models like LLAMA-2-70B (56.8\%) , WizardMath-70B (81.6\%), Gemini Pro (86.5\% with 32 trials) and GPT-3.5 (77.4\%). Most notably it can reach this level with only 200K examples (orders of magnitude less than other datasets).

\begin{table}[!htb]
\begin{center}
\small
\begin{tabular}{llrccc}
\Xhline{4\arrayrulewidth}
Model & Base model & Model size & Answer format & Eval method & GSM8K (\%)  \\
\Xhline{3\arrayrulewidth}
\multirow{4}{*}{Llama-2~\citep{llama2}} & \multirow{4}{*}{-}  & 7B &\multirow{3}{*}{nlp} & \multirow{3}{*}{pass@1} &14.6\\
& & 13B &&&  28.7\\
& & 34B &&&  42.2\\
& & 70B &&&  56.8\\
\hline
\multirow{3}{*}{MetaMath~\citep{metamath}} & \multirow{3}{*}{Llama-2} & 7B &\multirow{3}{*}{nlp}&\multirow{3}{*}{pass@1} &  66.5\\
& & 13B &&&  72.3\\
& & 70B &&&  \textbf{82.3}\\
\hline
\multirow{3}{*}{WizardMath~\citep{wizardmath}} & \multirow{3}{*}{Llama-2} & 7B & \multirow{3}{*}{nlp}&\multirow{3}{*}{pass@1} & 54.9\\
& & 13B &&&  63.9\\
& & 70B &&&  \textbf{81.6}\\
\hline
\multirow{4}{*}{MAmmoTH~\citep{mammoth}} & Code-Llama & 7B &  \multirow{3}{*}{code}&\multirow{4}{*}{pass@1} &  59.4\\
& Code-Llama & 12B &&&  64.7\\
& Code-Llama & 34B &&&  72.7\\
& Llama-2 & 70B & nlp &&  76.9\\
\hline
\multirow{2}{*}{Mistral~\citep{jiang2023mistral}} & \multirow{2}{*}{-} &7B & \multirow{2}{*}{nlp} & \multirow{2}{*}{maj1@8} &  52.2\\ 
&  & 8$\times$7B & & &  58.4\\
\hline
\multirow{2}{*}{OVM~\citep{OVM}} & Llama-2 & 7B+7B & \multirow{2}{*}{nlp}& \multirow{2}{*}{verify100@1} & 73.7\\
& Mistral & 7B+7B & &  & \textbf{84.7}\\
\hline
\multirow{2}{*}{Llemma~\citep{llemma}} & \multirow{2}{*}{Llama-2} & 7B & \multirow{2}{*}{nlp}&\multirow{2}{*}{pass@1} & 36.4\\
& & 34B & & & 51.5\\
\hline
\multirow{4}{*}{ToRA-Code~\citep{tora}} & \multirow{4}{*}{Llama-2} & 7B & \multirow{4}{*}{code} & \multirow{4}{*}{COT@1} & 72.6\\
& & 13B & & & 75.8\\
& & 34B & & & \textbf{80.7}\\
& & 70B & & & \textbf{84.3}\\
\hline
\multirow{2}{*}{Orca 2~\citep{mitra2023orca}} & \multirow{2}{*}{Llama-2} & 7B & \multirow{2}{*}{nlp} & \multirow{2}{*}{pass@1} & 55.72\\
& & 13B & & & 65.73\\

\Xhline{3\arrayrulewidth}
Gemini Pro & \multirow{2}{*}{-} & \multirow{2}{*}{-} & \multirow{2}{*}{nlp} & \multirow{2}{*}{maj1@32} & \textbf{86.5}\\
Gemini Ultra~\citep{google2023gemini} & & & & & 94.4\\
\midrule
GPT-3.5-0613 & \multirow{2}{*}{-} & \multirow{2}{*}{-} & \multirow{2}{*}{code} & \multirow{2}{*}{pass@1} & 77.4\\
GPT-4-0613~\citep{openai2023gpt4} & & & & & \textbf{97.0}\\
\Xhline{3\arrayrulewidth}
Phi-1.5~\citep{textbooks2} & - & 1.3B & code & \multirow{1}{*}{pass@1} & 44.6\\
\hline
\multirow{4}{*}{Phi-GSM \citep{liu2023tinygsm}} & Phi-1.5-tiny & 125M & \multirow{4}{*}{code} &  \multirow{4}{*}{pass@1}  & 63.1\\
& Phi-1.5-small & 350M & & & 65.9\\
& Phi-1.5 & 1.3B & & & 68.2\\
& Phi-2 & 2.7B & & & 74.3\\
\hline
\multirow{3}{*}{Phi-GSM+V \citep{liu2023tinygsm}} & Phi-1.5-tiny & 125M+125M & \multirow{3}{*}{code} &  \multirow{3}{*}{verify48@1}  & 68.9\\
& Phi-1.5-small & 350M+350M & & & 71.3\\
& Phi-1.5 & 1.3B+1.3B & & & \textbf{81.5}\\
\Xhline{4\arrayrulewidth}
Orca-Math & Mistral & 7B & nlp & \multirow{1}{*}{pass@1} & \textbf{86.81}\\
\hline
\end{tabular}
\end{center}
\caption{Results on GSM8K. The table is repurposed from \citep{liu2023tinygsm}. \textbf{Bold} labels indicate accuracies \textbf{exceeding 80\%}. The term '8$\times$7B' represents a blend of $8$ experts, with each expert having 7B parameters. '7B+7B' refers to the union of a 7B generation model and a 7B verification model. The addition of verifier models is signified by '+V'.} 
\label{tab:comparison}
\end{table}

\clearpage
\section{Dataset Construction: Agent-Instruct}
The goal of this step is to create a diverse set of grade school math word problems that contains both easy and hard problems. Towards this goal we create a variety of agents.

\paragraph{Seed Set}
We start by collecting sample math word problems from existing open-source datasets, namely NumGLUE \cite{mishra2022numglue}, AddSub \cite{hosseini2014learning}, ALGES \cite{kushman2014learning}, ASDiv \cite{miao2021diverse}, DRAW \cite{upadhyay2015draw}, GSM8k \cite{gsm8k}, MATHQA \cite{amini2019mathqa}, MultiArith \cite{roy2016solving}, SingeOP \cite{roy2015reasoning}, SingleEQ \cite{koncel2015parsing}, and SVAMP \cite{patel2021nlp}. We collect a total of $36,217$ problems. We utilize the Lila \cite{mishra2022lila} benchmark to collect the datasets. Specifically, we collect problems from the train and validation splits from Lila to construct the seed set. Interested readers, please refer to Lila \cite{mishra2022lila}.


\paragraph{Agent - Ask Me Anything}
We expand the seed set by creating multiple word problems from each problem in the seed set. We utilize the subsequent prompt for problem creation.

\begin{framed}
\noindent \textbf{Your goal is to create multiple word problems from a given word problem and its answer. First convert the question of the word problem into a statement. Then for each number in the converted problem create a new word problem. Here are some examples:}

\textbf{Example 1:}\
Q: Natalia sold clips to 48 of her friends in April, and then she sold half as many clips in May. How many clips did Natalia sell altogether in April and May?\

Answer: 72

\textit{Replacing question with statement}: Natalia sold clips to 48 of her friends in April, and then she sold half as many clips in May. Natalia sold altogether 72 clips in April and May.

\textbf{All questions:}\

\textless target\textgreater{} 48\

\textless question\textgreater{}
Natalia sold clips to some of her friends in April, and then she sold half as many clips in May. Natalia sold altogether 72 clips in April and May. How many clips did she sell in April?\

\textless target\textgreater{} half\

\textless question\textgreater{}
Natalia sold clips to 48 of her friends in April, and then she sold some clips in May. Natalia sold altogether 72 clips in April and May. What is the ratio of the number clips sold in April to number clips sold in May?\

\textless target\textgreater{} 72\

\textless question\textgreater{}
Natalia sold clips to 48 of her friends in April, and then she sold half as many clips in May. How many clips did Natalia sell altogether in April and May?

\textbf{Example 2:}\
Q: Weng earns \$12 an hour for babysitting. Yesterday, she just did 50 minutes of babysitting. How much did she earn?\

Answer: 10

\textit{Replacing question with statement}: Weng earns \$12 an hour for babysitting. Yesterday, she just did 50 minutes of babysitting. She earned \$10.

\textbf{All questions:}\

\textless target\textgreater{} 12\

\textless question\textgreater{}
Weng earns a certain amount per hour for babysitting. Yesterday, she just did 50 minutes of babysitting and earned 10. How much does she earn per hour?\

\textless target\textgreater{} 50\

\textless question\textgreater{}
Weng earns 12 an hour for babysitting. Yesterday, she just did some babysitting and earned 10. How much time did she spend on babysitting?\

\textless target\textgreater{} 10\

\textless question\textgreater{}
Weng earns 12 an hour for babysitting. Yesterday, she just did 50 minutes of babysitting. How much did she earn?

\textbf{Example 3:}\
Q: Betty is saving money for a new wallet which costs 100. Betty has only half of the money she needs. Her parents decided to give her 15 for that purpose, and her grandparents twice as much as her parents. How much more money does Betty need to buy the wallet?\

Answer: 5

\textit{Replacing question with statement}: Betty is saving money for a new wallet which costs 100. Betty has only half of the money she needs. Her parents decided to give her 15 for that purpose, and her grandparents gave her twice as much as her parents. She needs 5 more to buy the wallet.

\textbf{All questions:}\

\textless target\textgreater{} 100\

\textless question\textgreater{}
Betty is saving money for a new wallet. Betty has only half of the money she needs. Her parents decided to give her 15 for that purpose, and her grandparents twice as much as her parents. She needs 5 more to buy the wallet. What is the cost of the wallet?\

\textless target\textgreater{} half\

\textless question\textgreater{}
Betty is saving money for a new wallet which costs 100. She has some money saved, her parents decided to give her 15, and her grandparents gave her twice as much as her parents. Now, Betty needs 5 more to buy the wallet. What is the ratio of the money Betty have saved initially to the cost of wallet?\

\textless target\textgreater{} 15\

\textless question\textgreater{}
Betty is saving money for a new wallet which costs 100. She has half of the money she needs, her parents decided to give her some money, and her grandparents gave her twice as much as her parents. Now, Betty needs 5 more to buy the wallet. How much money did her parents give her?\

\textless target\textgreater{} twice\

\textless question\textgreater{}
Betty is saving money for a new wallet which costs 100. Betty has only half of the money she needs. Her parents decided to give her 15 for that purpose, and her grandparents also chipped in. Now, Betty needs 5 more to buy the wallet. What is the ratio of the amount given by her grandparents to the amount given by her parents?\

\textless target\textgreater{} 5\

\textless question\textgreater{}
Betty is saving money for a new wallet which costs 100. Betty has only half of the money she needs. Her parents decided to give her 15 for that purpose, and her grandparents twice as much as her parents. How much more money does Betty need to buy the wallet?

\textbf{Now solve this:}\

\textbf{Example 4:}\
Q: Your teacher is giving a test worth 200 points. There is a total of 30 5-point and 10-point questions. How many 5-point questions are on the test?\
Answer: 20
\end{framed}

Note that, the ``Ask Me Anything" agent is generating  problems based on the seed in Example 4. Examples 1 to 3 are provided as few-shot demonstrations. This agent creates a total of $120,445$ new problems. It is important to note that all the generated problems exhibit a similar narrative to that of the seed word problem. The solutions to these word problems are generated using GPT4-Trubo.

\paragraph{Agent - Suggester \& Editor}
We further expand the seed set by developing challenging problems. To achieve this, we introduce two new agents, namely \textit{Suggester} and \textit{Editor}. This setup involves the establishment of a team of agents working collaboratively to create a dataset geared toward a predefined objective. In this instance, the objective is to modify an existing problem to increase its difficulty.

The \textit{Suggester} examines a specific problem and proposes several methods for enhancing its complexity without creating the actual problem. Subsequently, the \textit{Editor} takes the original word problem and the Suggester's recommendations to generate an updated, more challenging problem. This iterative process can occur over multiple rounds, with each round further increasing the complexity of the previously generated problem. We utilize the AutoGen \citep{wu2023autogen} framework to implement the multi-agent workflow. Here is an example of the iterative process:

\begin{framed}
\noindent\textbf{Seed Problem:}\
Every hour Joanne has to collect the coins out of the fountain inside the mall. During the first hour, she collected 15 coins. For the next two hours, she collected 35 coins from the fountain. In the fourth hour, she collected 50 coins from the fountain but she gave 15 of them to her coworker so she could buy a soda. How many coins did she have after the fourth hour?

\noindent\textbf{Suggester:}\
Increase the number of hours, add more variables for each hour, and incorporate additional transactions.

\noindent\textbf{Editor:}\
Every hour, Joanne has to collect the coins from two separate fountains inside the mall - Fountain A and Fountain B. During the first hour, she collected 20 coins from Fountain A and 30 coins from Fountain B. In the next hour, she collected 25 coins more than she collected in Fountain A during the previous hour and 10 coins less than she collected in Fountain B during the previous hour. In the third hour, she collected the same number of coins in Fountain A as she collected in the second hour and additional 45 coins this time from Fountain B. During the fourth hour, she collected 90 coins in total, but she gave 20 coins to her coworker to buy a soda and 40 coins to another coworker to buy a meal. After the fifth hour, Joanne decided to donate 15

\noindent\textbf{Suggester:}\
Incorporate more fountains, increase the duration, and add more complex calculations for coins collected and spent or donated.

\noindent\textbf{Editor:}\
Joanne has to collect coins from four separate fountains inside the mall - Fountain A, Fountain B, Fountain C, and Fountain D every hour for 8 hours. In the first hour, she collects 20 coins from Fountain A, 30 coins from Fountain B, 40 coins from Fountain C, and 50 coins from Fountain D. In the subsequent hours, her collections have the following fluctuations:

\begin{itemize}
\item For Fountain A, she collects 10 more coins each hour than she collected in the previous hour.
\item For Fountain B, her collections decrease by 5 coins each hour compared to the previous hour.
\item For Fountain C, she collects double the number of coins each hour compared to the number of coins collected in Fountain A the same hour.
\item For Fountain D, the number of coins collected remains constant at 50 coins each hour.
\end{itemize}

At the end of the third hour, she donates 20

During the seventh hour, she has to share coins with her coworkers and splits them as follows: 15 coins for coworker one, 20 coins for coworker two, and she keeps 70

After 8 hours, Joanne decides to count her collected coins. How many coins did she have at the end of the eighth hour?
\end{framed}

We allow two rounds of iterations per problem and filter problems where the GPT4-turbo generated answer exceeds 1800 characters. At the end of this process, we collect $37,157$ problems.

\paragraph{DMath} Furthermore, we include $6,216$ problems sourced from DMath \citep{kim-etal-2023-aint}. These problems represent a subset of the $7,943$ problems present in the DMath training set, in which the solution computed by GPT4-Turbo aligns with the precise gold-standard answer.

\section{Training}
\subsection{Supervised Fine-Tuning Experiment (Iteration \#1)}
We finetune Mistral-7B on the Orca-Math-200K dataset. We have not used packing. The data is presented in the following instruction format:
\begin{framed}
    \begin{verbatim}
USER:\n{question}\n\nASSISTANT:\n{answer}
    \end{verbatim}
\end{framed}

The loss is computed only on the answer tokens. We employ a constant learning rate of $1\times10^{-6}$. The per-device batch size is set to $3$. Training is conducted for one epoch on eight A100 nodes, with each node containing eight GPUs.

\subsection{Iterative Learning from both Positive and Negative Signals}
\paragraph{Dataset Construction Iteration \#2}
To generate additional positive and negative solutions for each problem, we sample four responses from the SFT-tuned model from iteration \#1. Specifically, we utilize \textit{top\textunderscore p} $=0.95$ and \textit{temperature} $=0.7$. This process results in a dataset where each of the $200,000$ problems has one GPT4-Turbo generated solution and four student-generated solutions. Subsequently, we employ the prompt defined in GPT4-Based-Exact-Match (See section \ref{sec:eval} for details) to assess the alignment between the teacher's (GPT4-Turbo) answer and the student's answer. For all solutions where the student-generated answer does not match the teacher's answer, we label them as \textit{negative}; otherwise, we label the solution as \textit{positive}. We then construct the preference dataset as follows:
\begin{itemize}
    \item For each question, $q_i$ we construct $q_i^+$, the set of all \textit{positive} solutions for $q_i$. We treat the teacher solution as positive, thus this set by construction contains at least one element.
    \item For each question, $q_i$ we also construct $q_i^-$, the set of all \textit{negative} solutions for $q_i$. This set can be empty if all the $4$ responses are are aligned wrt the teacher's solution. Infact, this is the case for around $80k$ questions. For such situations, we randomly sample one response from $q_j^-$ for $4$ different $q_j$ where $j\ne i$ and $|q_j^-| > 0$. Note that, for this special situation $|q_i^+| = 4$.
    \item Let, $Q_i = \{(q_i, a_i^+, a_i^-)| (a_i^+, a_i^-) \in q_i^+ \times q_i^-\}$ be the preference dataset around $q_i$. The final preference dataset is created by taking the union of $Q_i$ for all $q_i$ in the training dataset.
\end{itemize}

\paragraph{Dataset Construction Iteration \#3}
Let M2 denote the model trained with \textit{KTO} \cite{ethayarajh2024kto} on the dataset constructed for Iteration \#2. We replicate the same procedure for the construction of dataset for Iteration \#3; however, we utilize M2 to generate the four responses instead of the SFT-tuned model from iteration \#1.

To learn from both positive and negative feedback, we have evaluated the performance of two algorithms: the Direct Preference Optimization (DPO) as described by \citep{rafailov2023direct} and the Kahneman-Tversky Optimization (KTO) introduced by \citep{ethayarajh2024kto}. DPO is a simple and popular approach for efficiently fine-tuning language models to align with  preferences. Additionally, we have explored the capabilities of KTO, which distinguishes itself by requiring only a binary ``yes'' or ``no'' response to assess the quality of an output. 

\section{Evaluation}
\label{sec:eval}
We use exact match as the metric. Given a model generated answer, we prompt GPT4 to extract the final short answer and match it with the gold short answer. We will refer to this metric as, \textit{GPT4-based-Exact-Match}. The following figure shows the prompt template:

\begin{framed}
\textbf{SYSTEM} \

As an expert Math teacher, your role is to evaluate a student's answer to a word problem. The problem is accompanied by a correct solution provided by the problem setter. It is important to remember that there may be various methods to solve a word problem, so the student's steps might not always align with those in the problem setter's solution. However, the final answer, typically a number, should be unique and match the problem setter's answer. Your task involves analyzing the student's solution to identify any mistakes and determine whether the answer can be modified to correct the error. If the student's answer is unfixable, consider creating practice problems to help improve their understanding.

Use the following format:

Error Analysis: \
In one sentence, extract the final answer from the problem setter's solution and compare it with the student's answer. Do they match?

Final Verdict: \
Correct/Incorrect

\textbf{USER} \

Question: \
Billy is volunteering his time to help people do their taxes. He can help 2 people per hour for 3 hours a day. If he takes 20\% of the days between March 1st and April 19th off, and helps people on all the other days. How many people does he help? (Remember there are 31 days in March.)

Problem Setter's answer: \

First find the number of days between March 1st and April 19th: 31 days + 19 days = 50 days \

Then subtract the percentage of time Billy takes off from 100\% to find the percentage of time he works: 100\% - 20\% = 80\% \

Then multiply that number by 80\% to find the number of days Billy works: 50 days * 80\% = 40 days \

Then find how many people Billy can help per day: 2 people/hour * 3 hours/day = 6 people/day \

Then multiply the number of people he helps per day by the number of days he works to find the total number of people he helps: 6 people/day * 40 days = 240 people

Student answer: \

First, let's calculate the total number of days between March 1st and April 19th. \

March has 31 days, and April has 30 days. So, the total number of days from March 1st to April 19th is: \

31 (March) + 19 (April) = 50 days \

Now, let's calculate how many days Billy takes off. He takes 20\% of the 50 days off. \

20\% of 50 days = 0.20 * 50 = 10 days \

So, Billy takes 10 days off. This leaves him with: \
50 days - 10 days = 40 days \

Billy helps people for 3 hours a day, and he can help 2 people per hour. So, in one day, he can help: \

3 hours/day * 2 people/hour = 6 people/day \

Now, let's calculate how many people he helps over the 40 days he works: \

40 days * 6 people/day = 240 people \

Therefore, Billy helps 240 people.

\textbf{ASSISTANT} \

Error Analysis: \
The student's final answer of helping 240 people matches the problem setter's solution.

Final Verdict: \
Correct
\end{framed}

\section{Results}
Table \ref{result_main} captures the performance of several training procedures on the GSM8k test set containing $1319$ word problems. We finetune Mistral-7B for upto three iterations. In first iteration we use supervised finetuning to obtain M1. For the second iteration we compare SFT, DPO \cite{rafailov2023direct} and KTO. The KTO trained model performs better in this group. We call this M2 and use M2 to generate the dataset for iteration \#3. For third iteration, we compare DPO and KTO where M2 servers as the starting point. We also compare these against three epochs of SFT training on the Orca-Math-200K dataset. For all SFT training  we employ a constant learning rate of $1\times10^{-6}$. The per-device batch size is set to $3$ and number-of-epochs is set to $1$. For DPO and KTO training jobs, we set beta to $0.3$, per-device batch size to $3$, gradient-accumulation-steps to $11$ and number-of-epochs $1$. For DPO and KTO training in iteration \#2 we employ a constant learning rate of $1\times10^{-6}$ and for iteration \#3 a constant learning rate of $1\times10^{-7}$.

\begin{table}[!htb]  
\centering  
\begin{tabular}{l|c}  
\hline  
Training Procedure & Pass@1 Accuracy on GSM8K Test set \\ \hline
SFT (M1) &  $79.91$ \\   
SFT (M1)  $\rightarrow$ SFT & $81.50$ \\   
SFT (M1) $\rightarrow$ DPO &  $84.23$\\   
SFT (M1) $\rightarrow$ KTO (M2) &  $85.06$\\   
SFT (M1) $\rightarrow$ SFT $\rightarrow$ SFT & $80.44$ \\ 
SFT $\rightarrow$ KTO (M2) $\rightarrow$ DPO & $84.91$\\
SFT $\rightarrow$ KTO (M2) $\rightarrow$ KTO (Orca-Math)& $\mathbf{86.87}$\\   
\hline
\end{tabular}
\vspace{5pt}
\caption{Table captures the performance of several iterative learning experiments and baselines on the GSM8k test set. SFT stands for one epoch of training on the Orca-Math-200K dataset. SFT $\rightarrow$ SFT stands two epochs of training on Orca-Math-200K. SFT $\rightarrow$ DPO (KTO) stands for one epoch of training on dataset for iteration \#2 with DPO (KTO) starting with M1. SFT $\rightarrow$ SFT $\rightarrow$ SFT stands for three epochs of training on Orca-Math-200K. SFT $\rightarrow$ KTO $\rightarrow$ DPO (KTO) stands for one epoch of training on dataset for iteration \#3 with DPO (KTO) starting with M2. For evaluation, we employ greedy decoding.} 
\label{result_main}  
\end{table}  

\subsection{Ablation Studies}
\paragraph{Model Generated Positives} We study the impact model generated \textit{positives} by limiting $q_i^+$ to contain only teacher generated solution. In other words we remove any $a_i^+$ that is model generated in the creation of the dataset for iteration \#2. Table \ref{result_ablation_positive} shows the result of training M1 with DPO and KTO on this dataset for one epoch. We reuse the hyperparameters for iteration \#2. Irrespective of the training algorithm, we see significant performance drop.
\begin{table}[h]  
\centering  
\begin{tabular}{l|l}  
\hline  
\hline 
M1 $\rightarrow$ DPO &  $81.96$ ($\textcolor{red}{-2.27}$)\\   
M1 $\rightarrow$ KTO & $82.79$ ($\textcolor{red}{-2.27}$)\\
\hline
\end{tabular}
\vspace{5pt}
\caption{Table captures that student generated \textit{positives} provide important supervision.}  
\label{result_ablation_positive}  
\end{table} 

\paragraph{Synthetic Negatives}
The preference dataset creation involves synthetic \textit{negative} creation in the situation where all four responses generated from M1 or M2 are \textit{positive}. We study the impact of these synthetic negatives by ignoring the questions, $q_i$, where all sampled responses are positive (Table \ref{result_ablation_negative}). This reduces the number of questions for iteration \#2 by around $80k$ and for iteration \#3 by around $104k$.

\begin{table}[h]  
\centering  
\begin{tabular}{l|l}  
\hline  
\hline 
M1 $\rightarrow$ DPO &  $60.73$ ($\textcolor{red}{-23.5}$)\\   
M1 $\rightarrow$ KTO & $85.22$ ($\textcolor{green}{+0.17}$)\\
M1 $\rightarrow$ KTO $\rightarrow$ KTO & $85.44$ ($\textcolor{red}{-1.43}$)\\
\hline
\end{tabular}
\vspace{5pt}
\caption{Table shows that the inclusion of problems where all sampled responses are positive is beneficial.}  
\label{result_ablation_negative}  
\end{table} 
\subsection{Math Benchmarks beyond GSM8k}
Table \ref{table:orca-math-other} presents the performance of Orca-Math on several other word problem datasets. For ease of evaluation, we selected datasets where the answer to each problem is a single number. The test sets of the benchmarks are obtained from Lila. We employ the GPT4-based exact-match metric, and model responses are generated using greedy decoding.

\begin{table}[h]  
\centering  
\begin{tabular}{|l|c|c|}  
\hline  
\textbf{Test Set} & \textbf{Orca-Math-Sft (M1)} &\textbf{Orca-Math} \\ \hline  
Addsub            & 88.99 &91.74              \\ \hline  
ASDiv             & 91.10 &91.10               \\ \hline  
MultiArith        & 98.28 &98.28              \\ \hline  
SingleOp          & 98.74 &99.37              \\ \hline  
SinglEq           & 97.25 &99.08              \\ \hline  
Svamp structured  & 87.63 &91.30               \\ \hline  
\end{tabular}
\vspace{5pt}
\caption{Performance of SFT trained model from Iteration \#1 (M1) and Orca-Math on AddSub, ASDiv, MultiArith, SingleOp, SinglEq, Svamp structured}  
\label{table:orca-math-other}  
\end{table}

\subsection{Contamination Check}  
We never use the test split of GSM8K or any other datasets during training or as seeds for synthetic problem generation. Nevertheless, We take the following approach for detecting any potential text contamination.
  
\begin{enumerate}  
    \item We begin by preprocessing the texts, which includes converting all characters to lowercase, removing punctuation, tokenizing the text into individual words, and removing common English stopwords to ensure uniformity in the data.  
    \item We then vectorize our text corpus using the Term Frequency-Inverse Document Frequency (TF-IDF) method and determine the cosine similarity between the test and training sets, from which we select the top-k (k=10) most analogous questions for each test query.  
    \item Finally, we evaluate the extent of text contamination by counting the number of test questions with the highest n-gram overlap above a preset threshold of 0.5 with their corresponding training set matches. We calculate the overlap of n-grams between pairs of texts using the Jaccard similarity. To conduct a rigorous contamination check, we set n=1. It is important to note that the n-gram overlap, when measured using Jaccard similarity, is a non-increasing function of n. 
    \item Upon executing our algorithm, we determined that the count of test questions exhibiting significant n-gram overlap is \textbf{eight}, thus indicating negligible text contamination within our test set according to the defined threshold. When limiting the train set to contain only the seed problems, the count of test questions exhibiting significant n-gram overlap is \textbf{seven}. Note that, for $n \geq 2$, the count of test questions exhibiting significant n-gram overlap is \textbf{zero}. 
    
\end{enumerate} 

\section{Related Works}
The generation of synthetic data through generative artificial intelligence (AI) models has evolved rapidly. Numerous datasets \cite{mitra2023orca, OpenOrca, mukherjee2023orca, wizardmath, ding2023enhancing, cui2023ultrafeedback, zhou2023lima, cheng2023adapting, wang2022self} have been proposed for both specialized and generic domains, with math-related datasets \cite{yu2023metamath, yue2023mammoth, templatemath2024, li2023camel} being closely related to our work.

Learning from rich signals has also garnered significant attention recently. Several studies \cite{rafailov2023direct, ethayarajh2024kto, liu2023statistical, azar2023general, chen2024self, yuan2024self}, have demonstrated the usefulness of preference learning. In this work, we present a detailed analysis of agent-based synthetic data generation and iterative preference learning in the grade school level math domain. Specifically, we demonstrate the robustness of KTO over DPO and the effectiveness of using model-generated positives to improve model training. We believe this is a preliminary step toward iterative learning and self improvement of small language models in challenging domains.

\section{Conclusions}
Our study provides compelling evidence that the mathematical reasoning capabilities of Small Language Models (SLMs) can be substantially enhanced. By employing iterative learning techniques and leveraging both positive and negative signals, we have successfully surpassed the previously perceived $80\%$ barrier on the GSM8k benchmark. Our 7B model, trained with 200K data, achieved an impressive $86.81\%$ accuracy. Furthermore, the incorporation of agents in dataset generation has proven to be a valuable approach, enabling the creation of more diverse and interesting datasets. These findings not only highlight the potential for significant improvements in SLM performance but also underscore the importance of innovative learning strategies and dataset generation methods in advancing the creation of powerful SLMs.

\bibliographystyle{plainnat} 
{
\small
\bibliography{anthology,custom}
}
\clearpage
\appendix

\end{document}